\documentclass{article}

\usepackage{arxiv}

\usepackage{authblk}

\usepackage{amsmath,amsfonts,amssymb}
\usepackage{graphicx}

\usepackage[ruled,vlined]{algorithm2e}
\SetKwComment{Comment}{$\triangleright$\ }{}

\usepackage{subfig}
\usepackage[dvipsnames]{xcolor}
\definecolor{bg_blue}{HTML}{0B3C5D}
\definecolor{bg_red}{HTML}{B82601}
\definecolor{bg_green}{HTML}{1C6B0A}
\definecolor{bg_light_blue}{HTML}{328CC1}
\definecolor{bg_light_grey}{HTML}{A8B6C1}
\definecolor{bg_yellow}{HTML}{D9B310}
\definecolor{bg_brown}{HTML}{6C5050}
\definecolor{bg_burgundy}{HTML}{76323F}
\definecolor{bg_olive_green}{HTML}{626E60}
\definecolor{bg_muted_olive}{HTML}{918770}
\definecolor{bg_beige}{HTML}{C09F80}
\definecolor{columbia_blue}{HTML}{B9D9EB}
\usepackage{url}
\usepackage[multiple]{footmisc}

\usepackage{xcolor}

\title{Dictionary-Learning-Based Data Pruning for System Identification}

\author[1,2]{Tingna Wang$^\dagger$}
\author[4]{Sikai Zhang$^\dagger$}
\author[1,3]{Mingming Song}
\author[1,2,3]{Limin Sun
\thanks{Corresponding author at: \texttt{tina\_wang@tongji.edu.cn}; \texttt{lmsun@tongji.edu.cn}}}

\affil[1]{College of Civil Engineering, Tongji University, Shanghai, China}
\affil[2]{Shanghai Qi Zhi Institute, Shanghai, China}
\affil[3]{State Key Laboratory of Disaster Reduction in Civil Engineering, Tongji University, Shanghai, China}
\affil[4]{Baosight Software, Shanghai, China}
\begin{document}
\maketitle
\def\thefootnote{$\dagger$}\footnotetext{Equal contribution}\def\thefootnote{\arabic{footnote}}

\begin{abstract}
System identification is normally involved in augmenting time series data by time shifting and nonlinearisation (e.g., polynomial basis), both of which introduce redundancy in features and samples.
Many research works focus on reducing redundancy feature-wise, while less attention is paid to sample-wise redundancy.
This paper proposes a novel data pruning method, called mini-batch FastCan, to reduce sample-wise redundancy based on dictionary learning.
Time series data is represented by some representative samples, called atoms, via dictionary learning.
The useful samples are selected based on their correlation with the atoms.
The method is tested on one simulated dataset and two benchmark datasets.
The R-squared between the coefficients of models trained on the full datasets and the coefficients of models trained on pruned datasets is adopted to evaluate the performance of data pruning methods.
It is found that the proposed method significantly outperforms the random pruning method.
\end{abstract}


\section{Introduction}
System identification refers to a method of identifying the mathematical description of a dynamic system using measured input-output data \cite{box2015time}. 
It can be used to forecast future values, assess the effects of input variations, design control schemes, etc. 
According to the objective of system identification,  it can be generally divided into two types \cite{billings2013nonlinear}. 
The first type focuses on the approximation scheme that produces the minimum production errors, such as fuzzy logic \cite{nelles785nonlinear} and neural networks \cite{miller1995neural}.
The second type focuses on the elucidation of the underlying rule that represents the system, such as spectral analysis \cite{jenkins1968spectral}, the Volterra series \cite{schetzen1980volterra} and Nonlinear Auto-Regressive with eXogenous inputs (NARX) \cite{chen1989representations}. 
NARX-based methods have gained increasing attention in various fields such as engineering, finance, biology, and social sciences due to their flexibility in modelling a variety of systems while maintaining interpretability \cite{kukreja2003narmax, billings2013nonlinear, boynton2018applications}.

To avoid model overfitting and reduce computational complexity, feature selection methods based on orthogonalisation techniques and greedy search are widely applied in constructing NARX models \cite{korenberg1988orthogonal, chen1989orthogonal, hong2008model}. 
The orthogonalisation-based method was firstly derived in \cite{korenberg1988orthogonal} to efficiently decide which terms should be included in the Nonlinear Autoregressive Moving Average with eXogenous input models. 
This method was then further developed in \cite{chen1989orthogonal} as orthogonal forward-regression estimators to identify parsimonious models of structure-unknown systems by modifying and augmenting some orthogonal least squares methods. 
A more comprehensive review of this feature selection idea and its development for system identification can be found in \cite{hong2008model}.

Similar to feature selection, input selection is also a crucial step in system identification, which can help to improve the identification performance and reduce storage and training costs \cite{goodwin1971optimal, mehra1974optimal, hong2008model}.
While optimal selection of input samples (often referred to as \textit{data pruning}) is currently an active research topic for computer vision \cite{raju2021accelerating, sorscher2022beyond, yang2024data} and natural language processing \cite{marion2023less, jin2024llm}, its application to system identification has received less attention, although it remains an intractable task. 
Data pruning plays a vital role in system identification for the following reasons. 
First, time-series data collected from continuous physical processes often exhibit strong temporal correlation, especially when sampled at high frequencies, resulting in redundant observations that hinder efficient model training. 
Second, the common use of time-shift operations to generate delayed signal versions further exacerbates data redundancy. 

This paper proposes a data-pruning method based on dictionary learning to select useful time series samples for constructing a concise mathematical model to represent a system accurately.
Specifically, the reduced polynomial NARX method is applied to the identification of nonlinear dynamic systems as an example. 
It is worth noting that the proposed method can also be used to select informative samples for other types of system identification models.
The canonical-correlation-based fast feature selection method introduced in \cite{zhang2022orthogonal, zhang2025canonical} is adopted to find the most important terms that should be included in the NAXR model to achieve the required accuracy. 
Dictionary learning is then creatively combined with the Fast selection method based on Canonical correlation (FastCan) to find the most useful time series samples to train these terms. 

Dictionary learning, initially popular in the signal processing area, is used to learn an overcomplete dictionary composed of signal atoms, where a sparse linear combination of these atoms can generate the original signal \cite{aharon2006k}.
It is currently widely studied for image processing since images usually admit sparse representation \cite{mairal2011task, vu2017fast}.
For the proposed method, k-means-based dictionary learning, introduced in Section \ref{sec:DPminiFC}, provides a dictionary of time terms, which is used as pseudo-labels in the data pruning process. 

\begin{figure}[htbp]
    \centering
    \includegraphics[width=1\linewidth]{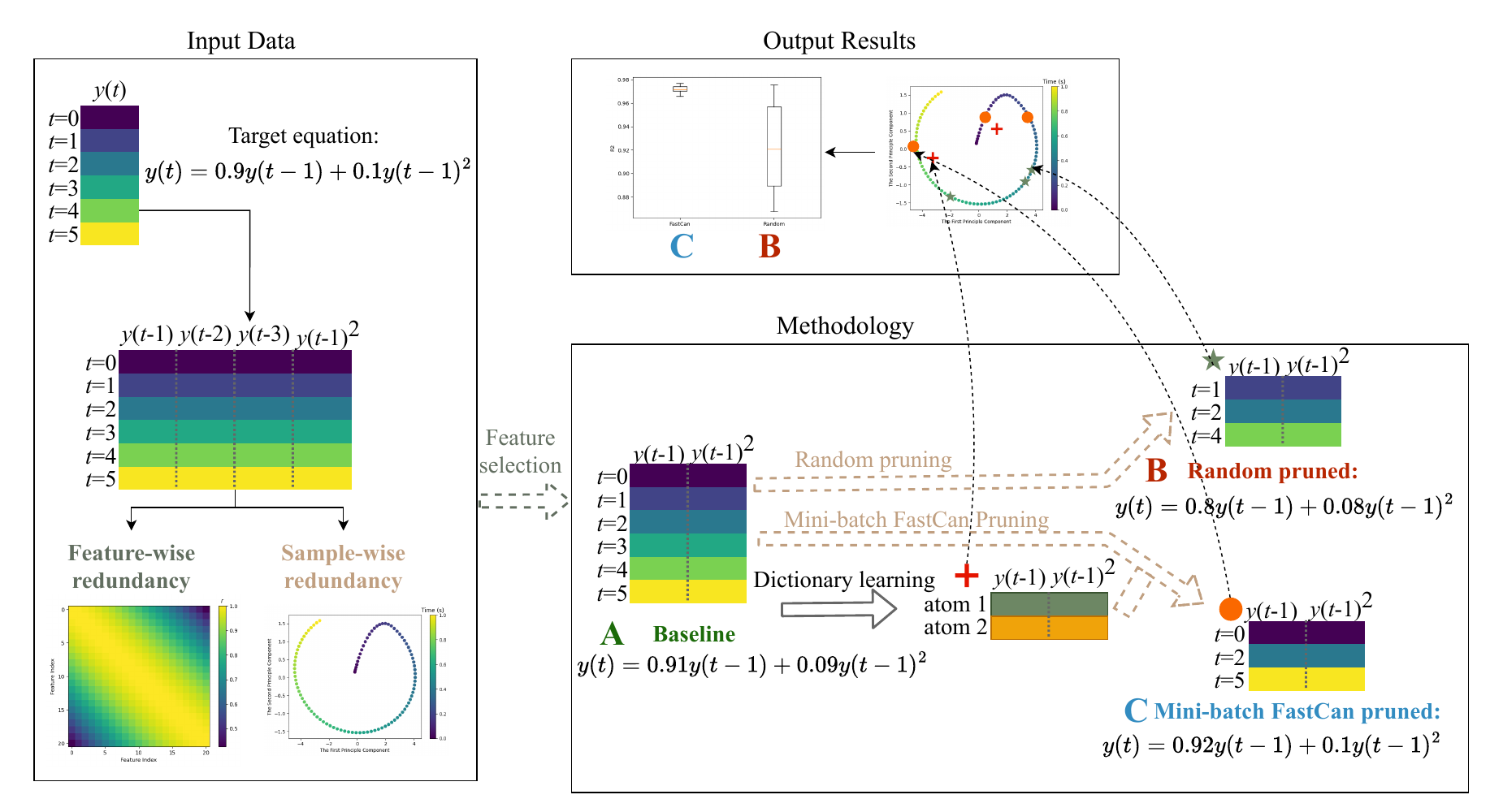}
    \caption[abs]{The schematic diagram of this research. The time series $y(t)$ is generated by the target equation, then time-shifted and nonlinearised to construct a redundant term library. 
    Feature-wise redundancy is represented by a correlation heatmap, while sample-wise redundancy is visualised by a principal component analysis (PCA) plot. 
    Feature selection is applied to the term library to reduce feature-wise redundancy. 
    \textbf{\textcolor{bg_green}{Dataset A}}\footnotemark, which contains the full set of samples, is used to train the \textbf{\textcolor{bg_green}{Baseline model}}. \textbf{\textcolor{bg_red}{Dataset B}}, a random subset of Dataset A, is used to train the \textbf{\textcolor{bg_red}{Random pruned model}}. 
    \textbf{\textcolor{bg_light_blue}{Dataset C}}, consisting of samples selected based on the atoms learned from Dataset A, is used to train the \textbf{\textcolor{bg_light_blue}{Mini-batch FastCan pruned model}}. 
    The selected samples obtained by both methods are visualised via a PCA plot.
    To compare the data pruning methods quantitatively, Dataset A is pruned repeatedly by random selection and mini-batch FastCan, and the models are then trained on the pruned datasets, respectively.
    Data pruning performance is then evaluated by the R-squared score between the coefficients learned from the pruned samples and the coefficients of the baseline model.}
    \label{fig:abs}
\end{figure}

In this paper, the proposed method, named \textit{mini-batch FastCan}, is compared with the random pruning method.
Random pruning commonly serves as a reliable benchmark in sample selection tasks and often yields better results than more advanced methods when 30\% or less of the data is retained \cite{ayed2023data}.
The reduced NARX trained with the full dataset is used as the baseline model.
The coefficients learned with the pruned dataset is compared with the coefficients of the baseline model.
The more the coefficients are close to the baseline model, the better the pruned dataset is.
The schematic diagram of the research in this paper is illustrated in Fig. \ref{fig:abs}.

The next section introduces the details of the proposed data-pruning method for system identification, followed by numerical case studies to demonstrate the necessity and advantages of the method. 
Subsequently, case studies on two public benchmark datasets \cite{janot02178709, schoukens2009wiener} for nonlinear system identification are presented, with a discussion on the effect of hyperparamteres in the proposed method. 

\footnotetext{The colour style used this figure is from \texttt{bg-mpl-stylesheets} \cite{bg-stylesheets}.}

\section{Methodology}
This section introduces the dictionary learning-based data pruning method, mini-batch FastCan, for a discrete system with finite orders. 

\subsection{Reduced Polynomial NARX}
The reduced polynomial NARX model is a nonlinear dynamic model that represents the system output as a sparse subset of polynomial terms consisting of past outputs and inputs selected from the full NARX model structure. It reduces complexity by including only important or relevant terms, rather than all possible combinations.
Mathematically, a NARX model is used to simulate a system, which is formulated as
\begin{equation}\label{eq:NARX}
\begin{split}
y(k) = & F[y(k-1),y(k-2),..., y(k-n_y),\\
       & u(k-1),u(k-2),..., u(k-n_u)]+e(k)
\end{split}
\end{equation}
where $y(k)$, $u(k)$ and $e(k)$ are the system output, input and noise sequences, respectively. $n_y$ and $n_u$ are the maximum lags for the system output and input. 
$F[\cdot]$ is a nonlinear function. 

The power-form polynomial model is adopted to approximate the nonlinear mapping $F[\cdot]$, and the equation (\ref{eq:NARX}) is then given as,
\begin{equation}\label{eq:NARX_Poly}
\begin{split}
y(k) = \theta_0 
&+ \sum_{i_1=1}^{n} f_{i_1}\big(x_{i_1}(k)\big) \\
&+ \sum_{i_1=1}^{n} \sum_{i_2=i_1}^{n} f_{i_1 i_2}\big(x_{i_1}(k), x_{i_2}(k)\big) 
+ \dots \\
&+ \sum_{i_1=1}^{n} \cdots \sum_{i_\ell=i_{\ell-1}}^{n} f_{i_1 i_2 \cdots i_\ell}\big(x_{i_1}(k), x_{i_2}(k), \dots, x_{i_\ell}(k)\big) 
+ e(k)
\end{split}
\end{equation}
where $\ell$ is the degree of polynomial nonlinearity, $n = n_y+n_u$, and 
\begin{align*}
f_{i_1 i_2 \cdots i_d}\big(x_{i_1}(k), x_{i_2}(k), \dots, x_{i_d}(k)\big) 
= \theta_{i_1 i_2 \cdots i_d} \prod_{j=1}^{d} x_{i_j}(k), \quad & 1 \leq d \leq \ell \\
x_{i_j}(k) =
\begin{cases} 
y(k-i), & 1 \leq i \leq n_y \\
u(k-i+n_y), & n_y+1 \leq i \leq n = n_y + n_u
\end{cases}
\end{align*}
where $\theta_{i_1 i_2 \cdots i_d}$ are model parameters, and $\prod_{j=1}^{d} x_{i_j}(k)$ are model terms whose order is not higher than $\ell$.

The total number of model terms in the polynomial NARX model given in equation (\ref{eq:NARX_Poly}) is $M = (n + \ell)!/[n! \, \ell!]$. 
It can be seen that the full NARX model can include a large number of terms, increasing the risk of overfitting.
Because only a subset of these terms is typically important to capture the underlying dynamic relationship \cite{billings2013nonlinear}, the canonical-correlation-based fast feature selection is carried out here to find out the $m$ significant model terms from $\prod_{j=1}^{d} x_{i_j}(k)$ with $y(k)$ as the target. 
Therefore, the reduced polynomial NARX model can be given as, 
\begin{equation*}\label{eq:NARX_Rdu_Poly}
y(k) = \theta_0 
+ \sum_{j=1}^{m} \sum_{i_1=1}^{n} \cdots \sum_{i_d=i_{d-1}}^{n} f_{i_1 i_2 \cdots i_d}\big(x_{i_1}(k), x_{i_2}(k), \dots, x_{i_d}(k)\big) + e(k), \quad 1 \leq m \leq M
\end{equation*}
Refer to \cite{zhang2022orthogonal, zhang2025canonical} for details on the specific feature selection steps. 

\subsection{Data Pruning with Mini-batch FastCan}\label{sec:DPminiFC}
For simplicity, the selected model terms are represented by the matrix $\mathbf{X}$ as follows,
\begin{equation*}
\mathbf{X} = 
\begin{pmatrix}
x_{1,1} & \dots & x_{1,N} \\
\vdots & \ddots & \vdots \\
x_{m,1} & \dots & x_{m,N} 
\end{pmatrix} 
\end{equation*}
where $m$ is the number of the features, i.e., the selected model terms, and $N$ is the number of time series samples given these selected features. 

A dictionary $\mathbf{D} \in \mathbb{R}^{m \times q}$ for the matrix $\mathbf{X} \in \mathbb{R}^{m \times N}$ is obtained by \textit{mini-batch k-means clustering} \cite{sculley2010web} and used as the target matrix for the subsequent data pruning step, where $q$ is the number of atoms in the dictionary, and typically $q > m$ for an overcomplete dictionary.
Each data sample $\mathbf{x}_{j} \in \mathbb{R}^m$ (column of $\mathbf{X}$) can be approximated as a sparse linear combination of the dictionary atoms, given by, 
\begin{equation*}
    \mathbf{x}_j \approx  \mathbf{D}\mathbf{a}_j
\end{equation*}
where $\mathbf{a}_j \in \mathbb{R}^q$ is a sparse coefficient vector for the $j$-th sample, which means most of the entries in $\mathbf{a}_j$  are zero.

\begin{algorithm}[t]
\caption{Mini-batch FastCan}\label{alg:mini_batch_DP}
\KwIn{$\mathbf{X} \in \mathbb{R}^{m \times N}$ \Comment*[r]{Sample matrix}\\
$q \in \mathbb{N}$ \Comment*[r]{Number of atoms in a dictionary $\mathbf{D}$}\\
$p \in \mathbb{N}$ \Comment*[r]{Batch size; Optional}\\
$n \in \mathbb{N}$ \Comment*[r]{Number of samples to select}}
\KwOut{$\mathbf{s} \in \mathbb{R}^{1\times n}$ \Comment*[r]{Selected indices}}

\textbf{Step 1:}\\
Apply the k-means-based dictionary learning  \cite{sculley2010web} to $\mathbf{X}^{\top}$ with $q$ clusters\;
The resulting $q$ cluster centres form the columns of a dictionary $\mathbf{D} \in \mathbb{R}^{m \times q}$ \Comment*[r]{Target matrix}
\textbf{Step 2:}\\
\If{$p$ is not specified \textbf{or} $p > \lceil n / q \rceil$}{
    Set $p \gets \lceil n / q \rceil$\;

    \If{$p >m $}{
        Set $p \gets m$\;
    }
}
\textbf{Step 3:}\\
Generate the batch matrix $\mathbf{B} \in \mathbb{R}^{q \times t}$, where $t =$ \( \lceil n/(q \times p) \rceil \), $\mathbf{B}[i, j]$ $ \leq p$ and $\sum_{i =1}^{q} \sum_{j =1}^{t} \mathbf{B}[i, j] = n$\;  
\textbf{Step 4:}\\
Initialize the candidate sample matrix $\mathbf{X}_c \gets \mathbf{X}$ and the target matrix $\mathbf{D}_c \gets \mathbf{D}$\;  

\For{$i \gets 1$ \textbf{to} $q$}{
    Let $\mathbf{d}_i \gets \mathbf{D}_c[:, i]$\;  
    \For{$j \gets 1$ \textbf{to} $t$}{
        Select $\mathbf{B}[i, j]$ samples from $\mathbf{X}_c$ by using the canonical-correlation-based fast selection method \cite{zhang2025canonical}, with $\mathbf{d}_i$ serving as the target vector\;  
        Append the indices of the selected samples to $\mathbf{s}$\;   
        Remove the selected samples from the candidate matrix $\mathbf{X}_c$\;  
    }
}
\Return $\mathbf{s}$\;
\end{algorithm}

The canonical-correlation-based fast selection method \cite{zhang2025canonical} is performed again, with $\mathbf{D}$ as the target, to find $n$ significant samples from the sample matrix $\mathbf{X}$. 
For selection methods which evaluate the linear association between candidates, no additional information is gained once the number of selected samples $n$ exceeds the rank of the data matrix $m$.
Therefore, to avoid invalid sample selection due to $n>m$, samples are selected in separate batches by the \textit{mini-batch FastCan} method, whose pseudocode is given in Algorithm \ref{alg:mini_batch_DP}, and the corresponding codes are available in the GitHub repository \footnote{\url{https://github.com/MatthewSZhang/data-pruning-sysid}}\footnote{\url{https://github.com/scikit-learn-contrib/fastcan}}.
Within each batch, the redundancy and interaction between samples are considered, while these are ignored between batches.

In Algorithm \ref{alg:mini_batch_DP}, if the batch size $p$ is not specified, it is set to $\lceil q/k \rceil$. 
If a given $p > \lceil n/q \rceil$, it is also reset to $\lceil q/k \rceil$, since exceeding this threshold will result in some atoms being excluded from the selection process. 
Additionally, this algorithm allows the number of atoms $q$ to be smaller than the number of features $m$, enabling a larger batch size to better capture sample redundancy.

\section{Numerical Case Studis}
\subsection{Visualisation of Sample Redundancy in System Identification}
\begin{figure}[htpb]
    \centering
    \includegraphics[width=0.47\linewidth]{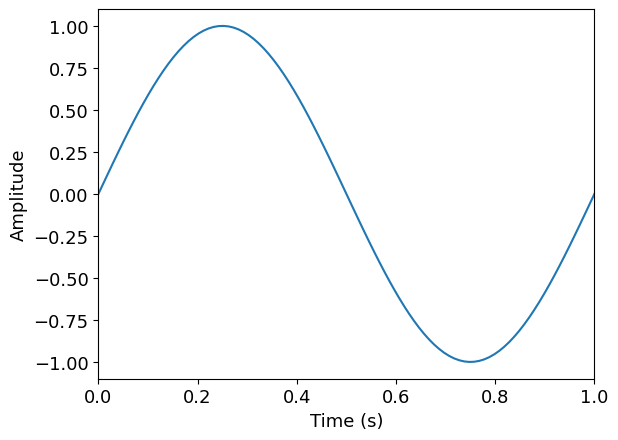}
    \caption{Time series of $y(t) = \text{sin}(2 \pi t)$ at $100$ Hz over $1 \text{s}$}
    \label{fig:run_raw}
\end{figure}

Here is an example to intuitively demonstrate the feature-wise and sample-wise redundancy that occurs in the system identification process.
A time series sampled from $y(t) = \text{sin}(2 \pi t)$ over a duration $1 \text{s}$ at a sample rate of $100$ Hz is illustrated in Fig. \ref{fig:run_raw}. 
After applying time shifting,  the one-dimensional data $y(t)$ is transformed into 20-dimensional data, as shown in Table \ref{tab:time_sft_data}, with a time step of $\Delta t = 0.01 \text{s}$.
There are 80 time series samples in total, where samples containing NaN values are excluded.

\begin{table}[htpb]
\caption{Time-shifted data representation with $\Delta t = 0.01 \, \text{s}$.}
\centering
\begin{tabular}{|c|c|c|c|c|c|}
\hline
$t$ (s)  & $y(t-\Delta t)$ & $y(t-2\Delta t)$ & $y(t-3\Delta t)$
& ... & $y(t-20\Delta t)$ \\ \hline
0                 & NaN      & NaN     & NaN      & ... & NaN    \\ \hline
0.01              & 0.000    & NaN     & NaN      & ... & NaN    \\ \hline
0.02              & 0.063    & 0.000   & NaN      & ... & NaN    \\ \hline
0.03              & 0.127    & 0.063   & 0.000   & ... & NaN     \\ \hline
0.04              & 0.189    & 0.127   & 0.063   & ... & NaN      \\ \hline
...               & ...      & ...     & ...     & ... & ...      \\ \hline
0.99              & -0.063   & -0.127  & -0.189   & ... & -0.955   \\ \hline
\end{tabular}
\label{tab:time_sft_data}
\end{table}

\begin{figure}[htpb]
\centering
    \subfloat[feature-wise]{
        \includegraphics[width=0.38\linewidth]{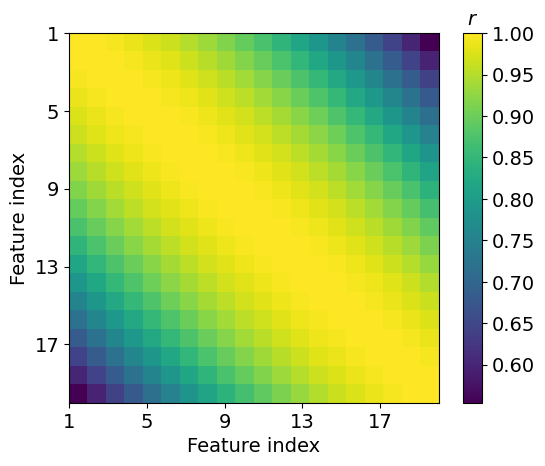}
        \label{fig:run_fea}
    }
    \hspace{1cm}
    \subfloat[sample-wise]{
        \includegraphics[width=0.41\linewidth]{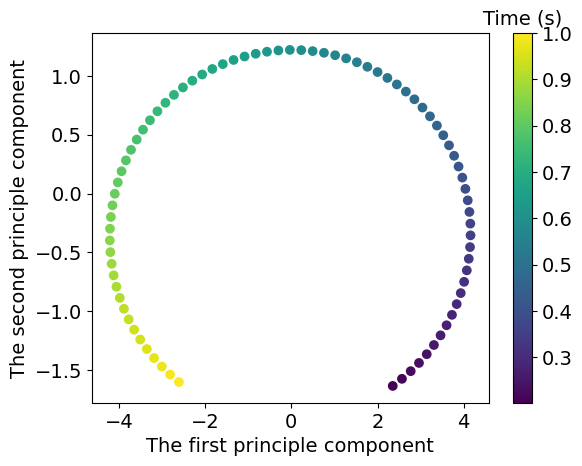}
        \label{fig:run_sam}
    }
    \caption{Correlation heatmap illustrating redundancy in data}
    \label{fig:run_data}
\end{figure}

The feature-wise redundancy is illustrated in the correlation heatmap of Fig. \ref{fig:run_fea}, where the $i^{th}$ feature corresponds to $y(t-i\Delta t)$ with $1 \leq i \leq 20$. 
The high Pearson's correlation near the main diagonal indicates redundancy between neighbouring features, such as $y(t-i\Delta t)$ and $y(t-(i+1)\Delta t)$.
This kind of redundancy can be mitigated via feature selection.

Fig. \ref{fig:run_sam} shows the redundancy within the time series samples using principal component analysis (PCA).
After projecting the 20-dimensional data into two-dimensional space, it is found that the data forms a continuous trajectory over time.
The proximity of adjacent samples in the lower-dimensional space given by PCA indicates that there is redundancy between these time series samples. 
This redundancy can be addressed by data pruning.

\subsection{Data with Dual Stable Equilibria}
\begin{figure}[htbp]
    \centering
    \includegraphics[width=0.45\linewidth]{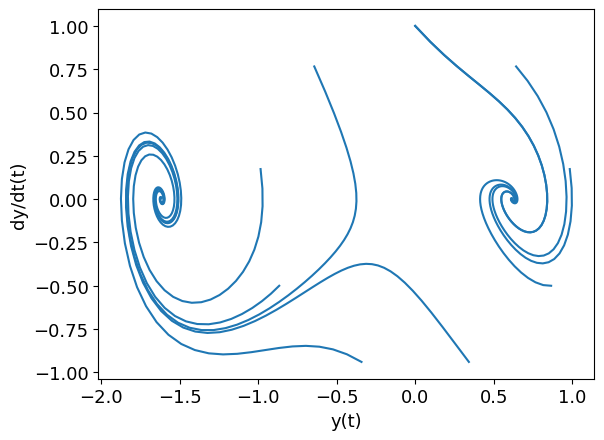}
    \caption{A phase portrait of the nonlinear system with dual stable equilibria}
    \label{fig:pp_dsed}
\end{figure}

The simulated time series data is generated by a non-autonomous nonlinear system, given by,
\begin{equation*}
\ddot{y} + \dot{y} - y + y^2 + y^3 = u
\end{equation*}
where $u(t) = 0.1 \cos (0.2 \pi t)$. 
By setting $u(t)=0$ and initialising $y$ and $\dot{y}$ with different values, a phase portrait of the nonlinear system is obtained, as shown in Fig. \ref{fig:pp_dsed}.

It can be seen that this system exhibits two stable equilibria, with five measurements for each of the left and right equilibria. 
Therefore, this dataset is referred to as \textit{symmetrical dual-stable-equilibria} (SDSE) dataset.
To comprehensively capture its dynamics, the data pruning algorithm should identify two distinct kinds of samples attracted to the different stable spirals and select samples from both kinds.
Unlike feature selection, random selection serves as a strong baseline for sample selection and often outperforms more sophisticated algorithms when retaining 30\% or less of the data \cite{ayed2023data}.
Therefore, the results of the proposed data-pruning method will be compared with those of the random selection method to demonstrate its advantages.

\begin{figure}[htpb]
\centering
    \subfloat[SDSE dataset]{%
        \includegraphics[width=0.40\linewidth]{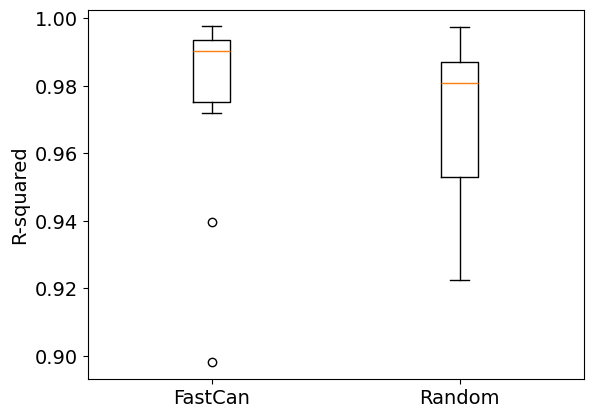}
        \label{fig:box_sdsed}%
    }
    \hspace{1cm}
    \subfloat[ADSE dataset]{%
        \includegraphics[width=0.40\linewidth]{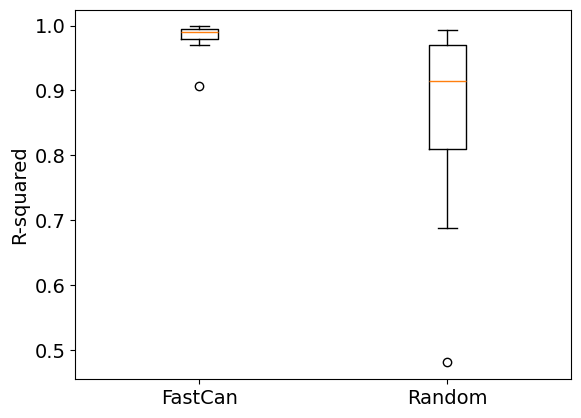}
        \label{fig:box_adsed}%
    }
    \caption{Comparison of sample selection performance between mini-batch FastCan and random selection methods using R-squared on two datasets}
    \label{fig:box_dsed}
\end{figure}

To quantitatively evaluate the selection performance of the mini-batch FastCan and random methods, a reduced polynomial NARX model with ten terms and an intercept is derived using the full dataset as the baseline.
The model coefficients trained on the full training dataset are used as the baseline results.
A more detailed description of the baseline NARX model and its prediction performance on the test dataset are provided in Appendix 1. 
The NARX terms are then fixed, and the model coefficients are trained using the pruned training datasets obtained via the mini-batch FastCan and random selection methods. 
100 samples are then selected as an example, and the corresponding number of atoms for dictionary learning is set to 15.
The reasons for this number of atoms will be discussed in Section \ref{sec:eff_hypers}.
The coefficients trained on the two pruned datasets are compared with those trained on the full dataset using the R-squared. 
The sample selection and training processes are repeated ten times, with the results presented in Fig. \ref{fig:box_sdsed}. 

\begin{figure}[htpb]
\centering
    \subfloat[SDSE dataset]{
    \includegraphics[width=0.40\linewidth]{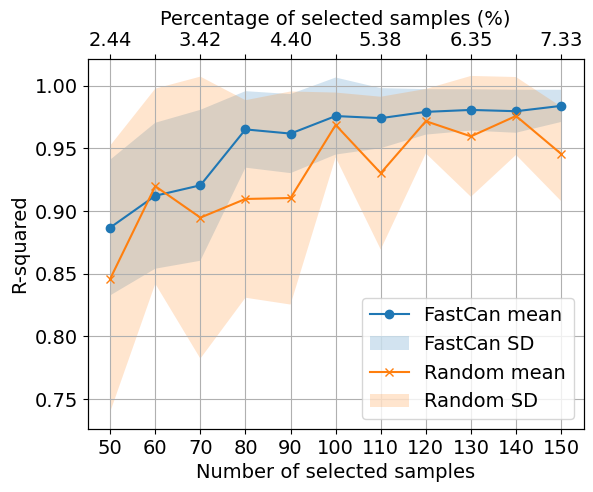}
    \label{fig:errbar_sdsed}
    }
    \hspace{1cm}
    \subfloat[ADSE dataset]{
    \includegraphics[width=0.40\linewidth]{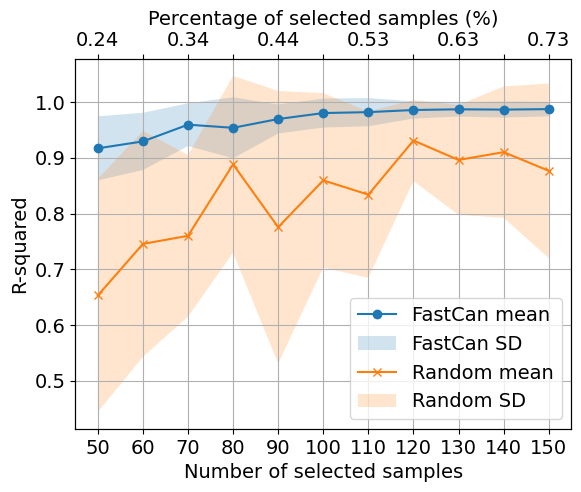}
    \label{fig:errbar_adsed}
    }
    \caption{The effect of sample size on the performance of mini-batch FastCan and random selection methods for two datasets. SD stands for standard deviation.}
    \label{fig:errbar_dsed}
\end{figure}

As shown in Fig. \ref{fig:box_sdsed}, the mini-batch FastCan method gives results with a higher median and lower variance, indicating superior and more consistent performance compared to random selection.
As illustrated in Fig. \ref{fig:errbar_sdsed}, the same pattern is observed when the number of selected samples varies between 50 and 150 and the number of atoms is fixed to 15, except for 60 samples---this case will be discussed in Section \ref{sec:eff_hypers}.
Furthermore, for the mini-batch FastCan method, the variance of the model coefficients decreases as the number of selected samples increases. However, this relationship does not exist when the samples are randomly selected in Fig. \ref{fig:errbar_sdsed}.

\begin{figure}[htpb]
\centering
    \subfloat[SDSE dataset]{
        \includegraphics[width=0.40\linewidth]{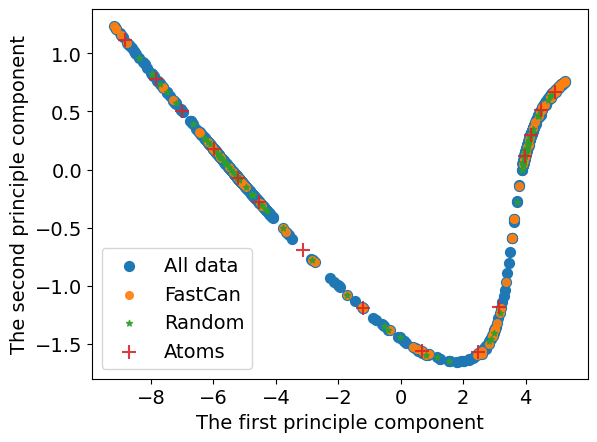}
        \label{fig:pca_sdsed}

    }
    \hspace{1cm}
    \subfloat[ADSE dataset]{
        \includegraphics[width=0.40\linewidth]{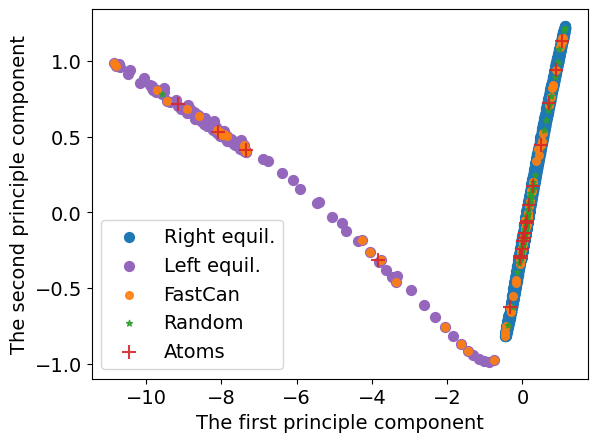}
        \label{fig:pca_adsed}
    }    
    \caption{Visualisation of the data structure of two datasets}
    \label{fig:pca_dsed}
\end{figure}

To examine whether the sample distributions corresponding to the results shown in Fig. \ref{fig:box_sdsed} are different, the structure of the SDSE data is visualised using its first and second principal components, as shown in Fig. \ref{fig:pca_sdsed}.
The selected samples corresponding to the results shown in Fig. \ref{fig:box_sdsed} are projected onto these principal directions, with their distribution shown in Fig. \ref{fig:pca_sdsed}.
It can be observed that the samples selected via mini-batch FastCan tend to cluster around the learned atoms, and the randomly selected samples are evenly distributed across the entire dataset according to the spatial distribution of candidate samples.
Compared to random selection,  the proposed method makes more samples distributed at the left and right ends.
Nevertheless, the overall difference between the two sample distributions is relatively small.

To further determine whether the sample distribution selected by the proposed method is different from that of the random method, 2 measurements in left equilibrium and 98 measurements in right equilibrium are simulated to repeat the previous analysis. 
This dataset is referred to as \textit{asymmetrical dual-stable-equilibria} (ADSE) dataset. 
100 samples are selected, and the number of atoms is set to 20 (see Section \ref{sec:eff_hypers} for details on this setting).
The results are presented in Figs. \ref{fig:box_adsed}-\ref{fig:pca_adsed}.
As shown in Figs. \ref{fig:box_adsed} and \ref{fig:errbar_adsed}, when the number of samples across different conditions is imbalanced, the mini-batch FastCan method has obvious advantages over the random sample selection method.
Furthermore, as shown in Fig. \ref{fig:pca_adsed}, almost all samples selected by the random method are concentrated in the right area, that is, these samples belong to the right equilibrium. 
In contrast, the proposed method selects samples from both equilibria, with those from the left equilibrium evenly distributed in the left area. 
These phenomena indicate that, unlike random selection, the mini-batch FastCan method is less influenced by the imbalance in the candidate sample distribution and is more effective at capturing diverse system behaviours.
This phenomenon, in turn, helps explain why the identification performance corresponding to the proposed method is superior to that corresponding to the random selection method.

\section{Case Studies on the Benchmark Datasets}
Two benchmark datasets, collected from the Electro-Mechanical Positioning System (EMPS) \cite{janot02178709} and the Wiener-Hammerstein System (WHS) \cite{schoukens2009wiener} are adopted to evaluate the performance of the mini-batch FastCan method for data pruning in real-world system identification tasks. 

\subsection{Data from the Electro-Mechanical Positioning System}
The EMPS is a standard drive configuration for prismatic joints of robots or machine tools. The primary nonlinearity of the corresponding data is introduced by friction effects. 
The baseline NARX model with ten terms and an intercept is derived, and the model coefficients trained on the full training dataset are used as the baseline to quantitatively evaluate the selection performance of the mini-batch FastCan and random methods.
See Appendix 1 for the performance of the baseline NARX on the test dataset.

\begin{figure}[htpb]
\centering
    \subfloat[EMPS dataset]{
        \includegraphics[width=0.40\linewidth]{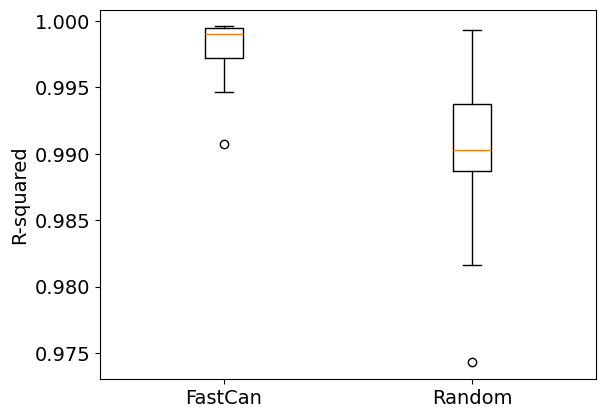}
        \label{fig:box_emps}
    }
    \hspace{1cm}
    \subfloat[WHS dataset]{
        \includegraphics[width=0.40\linewidth]{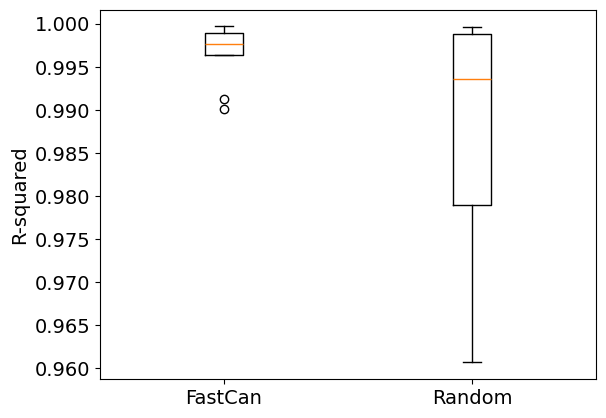}
        \label{fig:box_whbm}
    }
    \caption{Comparison of sample selection performance between mini-batch FastCan and random selection methods using R-squared on two benchmark datasets}
    \label{fig:box_benchmark}
\end{figure}

\begin{figure}[htpb]
\centering
    \subfloat[EMPS dataset]{
        \includegraphics[width=0.40\linewidth]{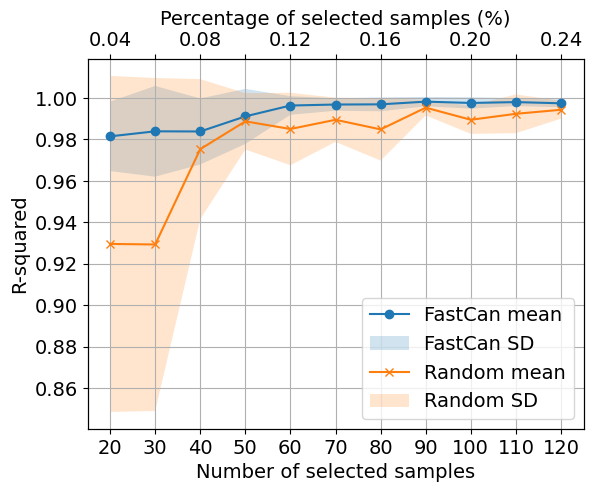}
        \label{fig:errbar_emps}
    }
    \hspace{1cm}
    \subfloat[WHS dataset]{
        \includegraphics[width=0.40\linewidth]{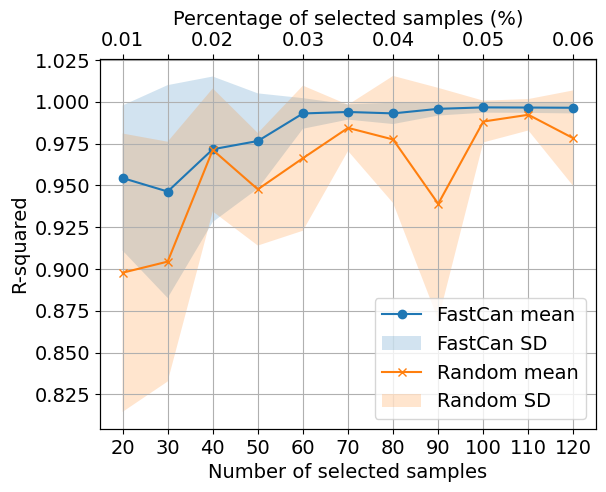}
        \label{fig:errbar_whbm}
    }
    \caption{The effect of sample size on the performance of mini-batch FastCan and random selection methods for two benchmark datasets. SD stands for standard deviation.}
    \label{fig:errbar_benchmark}
\end{figure}

The number of selected samples is 100, and the number of atoms for dictionary learning is set to 25 (see Section \ref{sec:eff_hypers} for details on this setting).
The coefficients trained on the two pruned datasets are compared with those from the full dataset using the R-squared, with the NARX terms fixed.
The sample selection and training processes are repeated ten times, with the results illustrated in Fig. \ref{fig:box_emps}. 
In addition, Fig. \ref{fig:errbar_emps} presents the results corresponding to two methods for different numbers of selected samples, ranging from 20 to 120.

Consistent with the observations from the SDSE and ADSE datasets, the mini-batch FastCan method produces results with a higher median and lower variance, indicating more stable and reliable performance.
Additionally, as expected, the variance of the model coefficients also decreases as the number of selected samples increases for the mini-batch FastCan method.

\begin{figure}[htpb]
\centering
    \subfloat[EMPS dataset]{
        \includegraphics[width=0.40\linewidth]{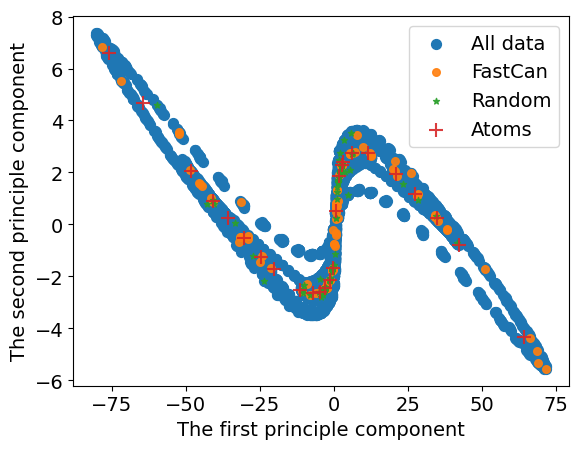}
        \label{fig:pca_emps}
    }
    \hspace{1cm}
    \subfloat[WHS dataset]{
        \includegraphics[width=0.40\linewidth]{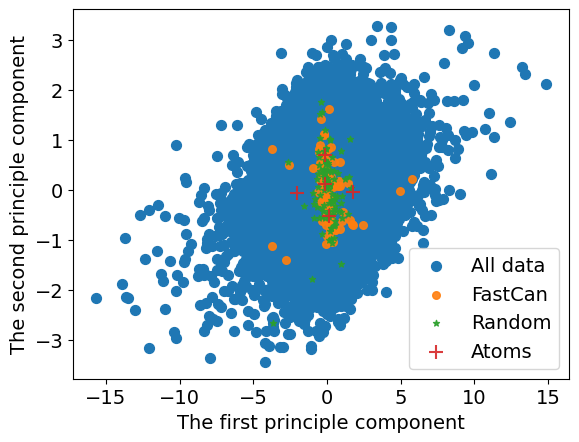}
        \label{fig:pca_whbm}
    }
    \caption{Visualisation of the data structure of two benchmark datasets}
    \label{fig:pca_benchmark}
\end{figure}

The structure of the data collected from the EMPS is visualised by its first two principal components, as shown in Fig. \ref{fig:pca_emps}. 
The atoms determined for the dictionary learning and the selected samples corresponding to the results given in Fig. \ref{fig:box_emps} are projected onto this reduced space.
The atoms and selected samples corresponding to the mini-batch FastCan method are distributed in the central, left and right regions, while the randomly selected samples are almost only spread over the central part.

\subsection{Data from the Wiener-Hammerstein System}
The WHS is a well-known block-oriented structure consisting of a static nonlinearity between two linear time-invariant blocks.
The same strategy is applied to quantitatively evaluate the sample selection performance of the mini-batch FastCan and random methods for the identification task of WHS. 
The baseline NARX model with ten terms and an intercept is derived, and its performance on the test dataset is provided in Appendix 1.

The number of atoms for dictionary learning is set to 5 (see Section \ref{sec:eff_hypers} for details on this setting, and 100 samples are selected, with the R-squared results shown in Fig. \ref{fig:box_whbm}.
Then, the number of selected samples varies between 20 and 120 to demonstrate the variation of R-squared with the selected sample size, as presented in Fig. \ref{fig:errbar_whbm}.
Fig. \ref{fig:box_whbm} demonstrates that the results obtained by the mini-batch FastCan are better than those obtained by the random method. 
This phenomenon is also observed in Fig. \ref{fig:errbar_whbm} except at 40 samples.
This exception will be analysed and further discussed in the following section.

The structure of the data collected from the WHS is visualised in Fig. \ref{fig:pca_whbm} using its first two principal components.
The atoms determined for the dictionary learning and the selected samples are shown in Fig. \ref{fig:pca_whbm}, with the number of atoms set to 5.
Different from the observations from the ADSE data and the EMPS data, the samples obtained by the mini-batch FastCan method and the random selection method have similar distributions, with the selected samples mainly distributed over the central part of the dataset.
This phenomenon is unexpected because the samples associated with the results in Fig. \ref{fig:box_whbm} exhibit similar distributions but provide varying levels of information for system identification. 
This observation suggests that sample selection strategies based solely on distributional characteristics of the data space may be insufficient for identifying informative samples.

\begin{figure} [t]
\centering
    \subfloat[SDSE dataset]{%
      \includegraphics[width=0.40\columnwidth]{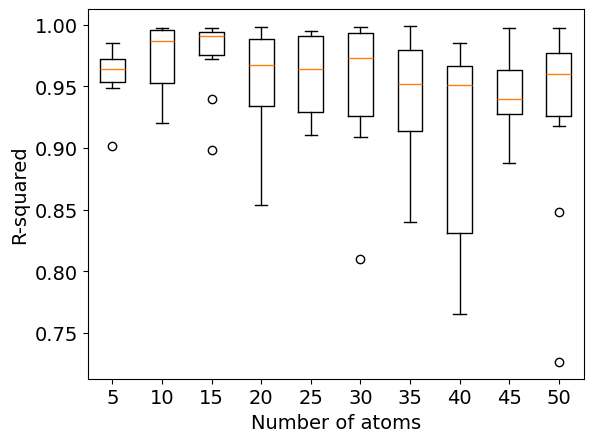}
      \label{fig:atom_dsed}%
    }
    \hspace{1cm}
    \subfloat[ADSE dataset]{%
      \includegraphics[width=0.40\columnwidth]{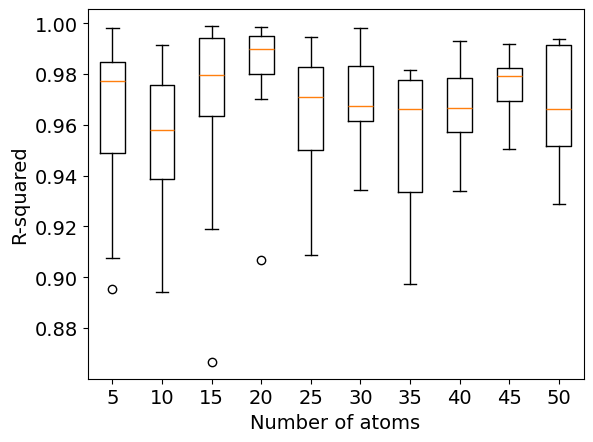}
      \label{fig:atom_dsed_eq}%
    }
    \vspace{0.25cm}
    \subfloat[EMPS dataset]{%
      \includegraphics[width=0.40\columnwidth]{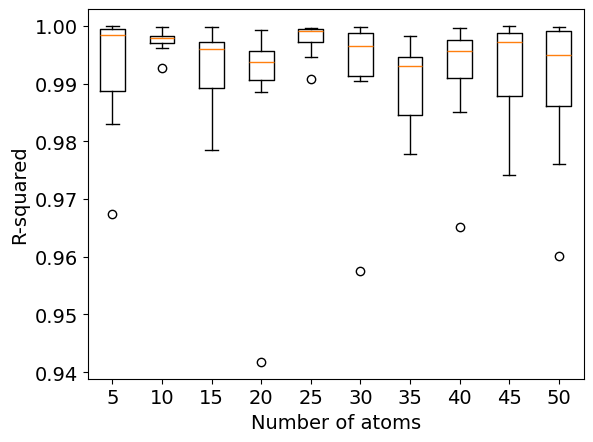}
      \label{fig:atom_emps}%
    }
    \hspace{1cm}
    \subfloat[WHS dataset]{%
      \includegraphics[width=0.40\columnwidth]{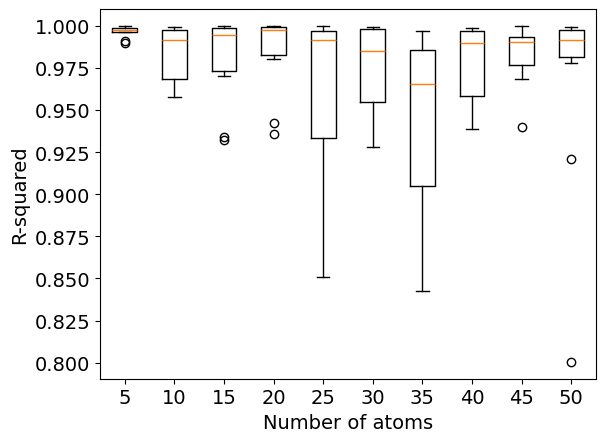}
      \label{fig:atom_whbm}%
    }
    \caption{The effect of atom size on the performance of mini-batch FastCan method}\label{fig:atom_four}
\end{figure}

\section{The Effect of Hyperparameters in the Mini-batch FastCan Method}\label{sec:eff_hypers}
\begin{figure} [htbp]
\centering
    \subfloat[SDSE dataset]{%
      \includegraphics[width=0.40\columnwidth]{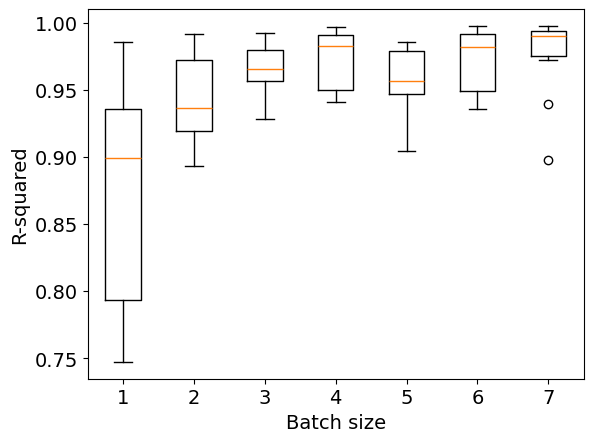}
      \label{fig:batch_dsed}%
    }
    \hspace{1cm}
    \subfloat[ADSE dataset]{%
      \includegraphics[width=0.40\columnwidth]{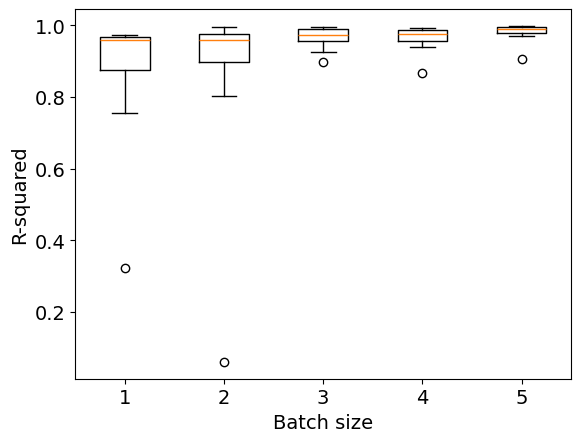}
      \label{fig:batch_dsed_eq}%
    }
    \vspace{0.25cm}
    \subfloat[EMPS dataset]{%
      \includegraphics[width=0.40\columnwidth]{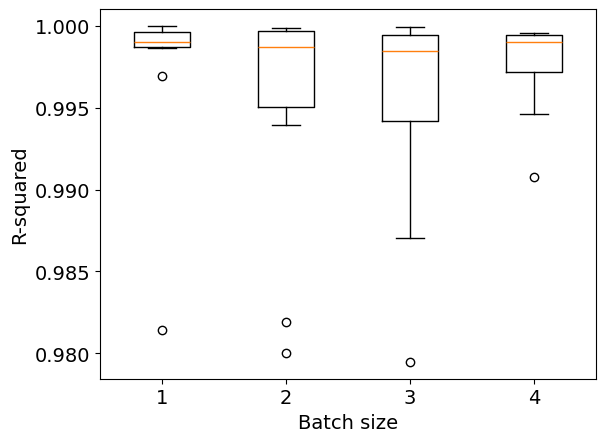}
      \label{fig:batch_emps}%
    }
    \hspace{1cm}
    \subfloat[WHS dataset]{%
      \includegraphics[width=0.40\columnwidth]{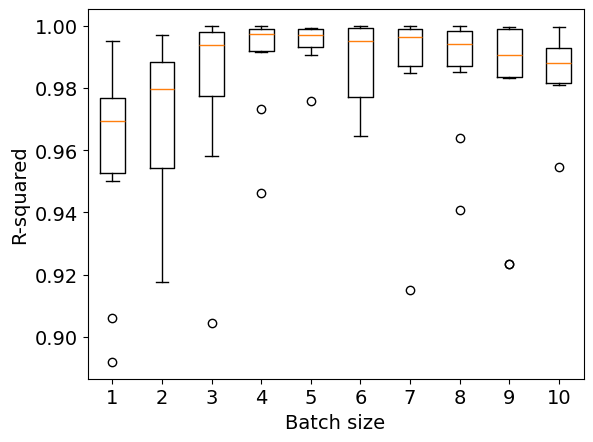}
      \label{fig:batch_whbm}%
    }
    \caption{The effect of batch size on the performance of mini-batch FastCan method under the optimal atom size}\label{fig:batch_four}
\end{figure}

In the mini-batch FastCan method, sample selection results are influenced by two hyperparameters: the number of atoms in the dictionary and the batch size. 
To examine how these two hyperparameters affect the performance of the selected samples, system identification results against the varying atom size and batch size for the four previously used datasets are given in Fig. \ref{fig:atom_four} to Fig. \ref{fig:batch_four}.
For all datasets, $100$ samples are selected.
For the results in Fig. \ref{fig:atom_four}, the batch size is not given as an input parameter but is instead set to the ceiling of the ratio between the number of selected samples and the number of atoms (\( \lceil n/q \rceil \)), with the aim of minimising redundancy among the selected samples when all atoms are used in the sample selection process. Therefore, the batch size varies for each atom size in Fig. \ref{fig:atom_four}.

As shown in Fig. \ref{fig:atom_four}, the performance deteriorates when the atom size exceeds the size associated with the highest R-squared value. 
This may be attributed to the fact that, for a fixed number of selected samples, larger atom sizes lead to smaller batch sizes, thus increasing sample-wise redundancy.
In practical engineering applications, the optimal atom size can be determined by defining a candidate range and selecting the value that yields the best performance via an exhaustive search.

\begin{figure} [htbp]
\centering
    \subfloat[SDSE dataset]{%
      \includegraphics[width=0.45\columnwidth]{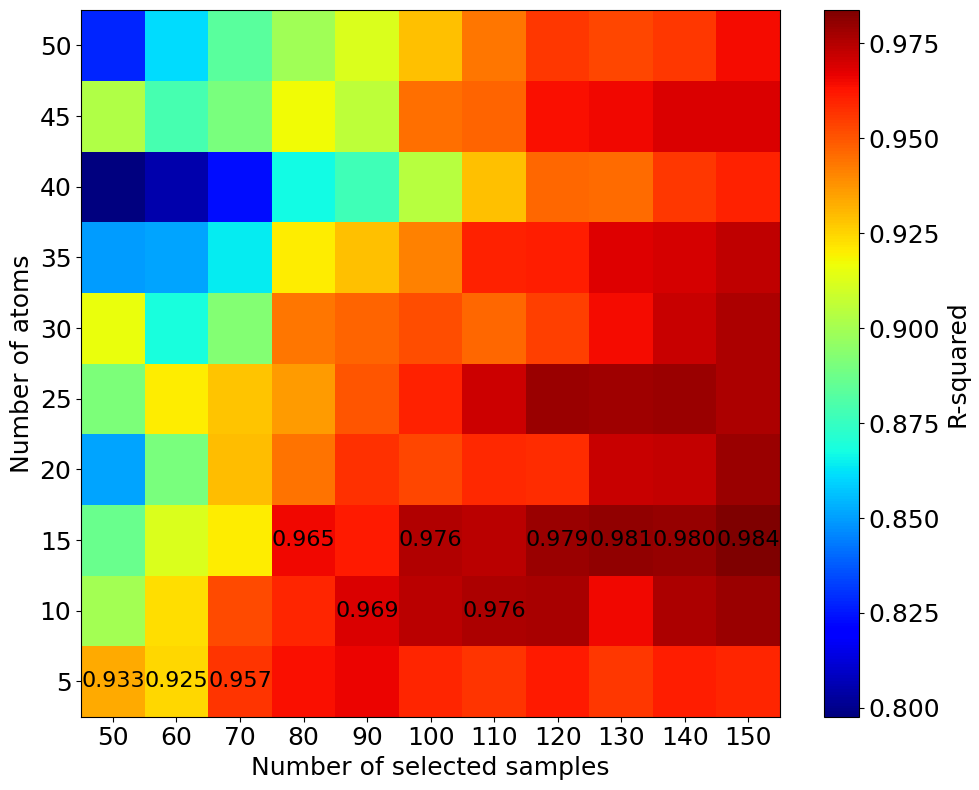}
      \label{fig:sample_dsed}%
    }
    \hspace{0.3cm}
    \subfloat[ADSE dataset]{%
      \includegraphics[width=0.45\columnwidth]{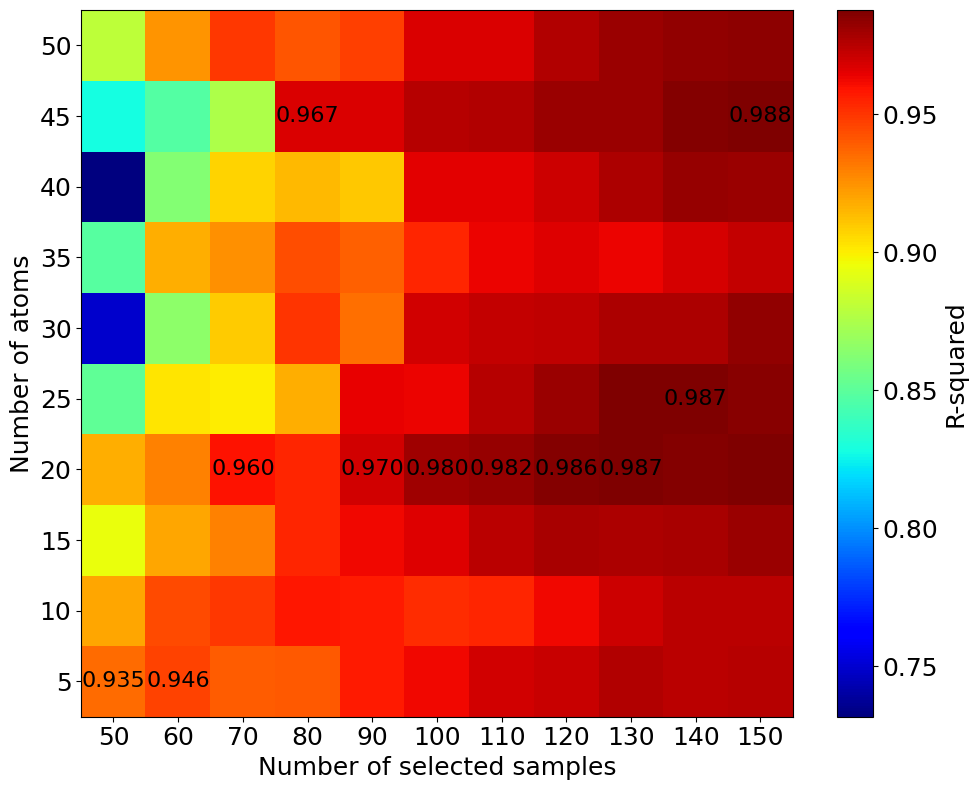}
      \label{fig:sample_dsed_eq}%
    }
    \vspace{0.25cm}
    \subfloat[EMPS dataset]{%
      \includegraphics[width=0.45\columnwidth]{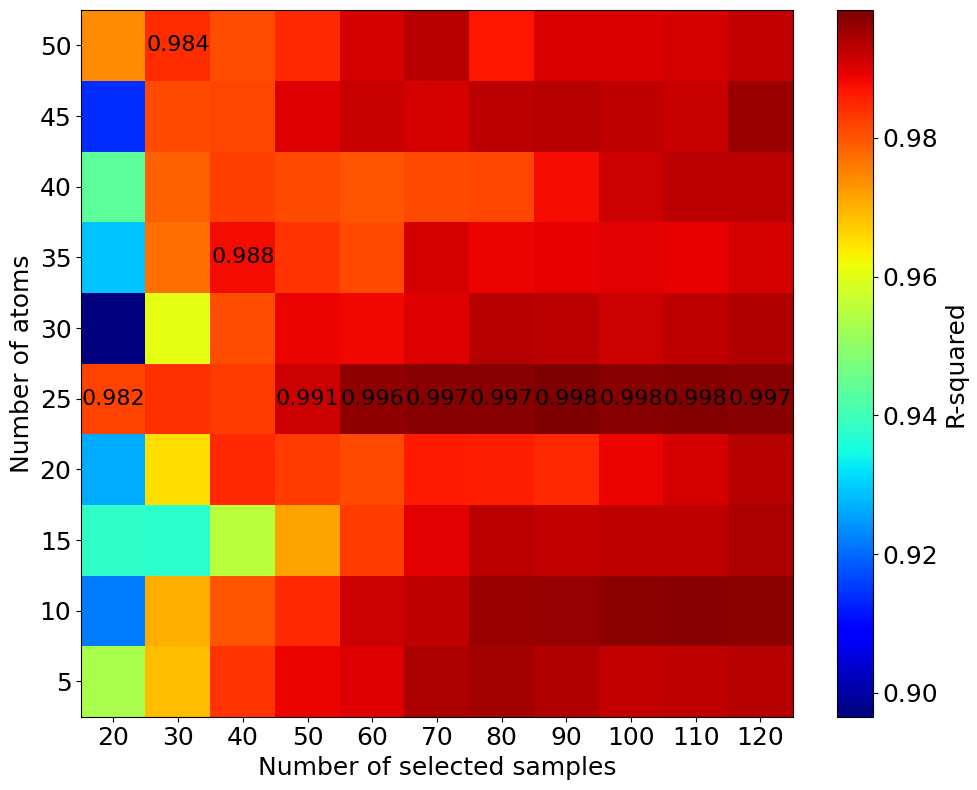}
      \label{fig:sample_emps}%
    }
    \hspace{0.3cm}
    \subfloat[WHS dataset]{%
      \includegraphics[width=0.45\columnwidth]{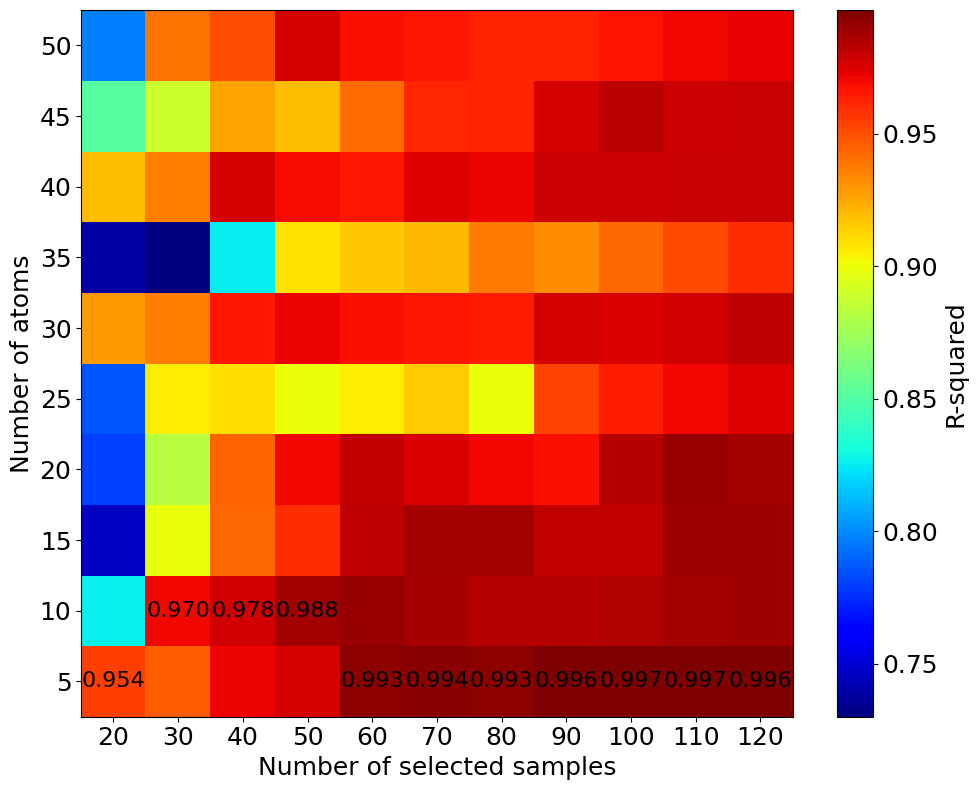}
      \label{fig:sample_whbm}%
    }
    \caption{The effect of atom size on the performance of mini-batch FastCan method under different sample numbers}\label{fig:sample_four}
\end{figure}

\begin{figure} [htbp]
\centering
    \subfloat[SDSE dataset]{%
      \includegraphics[width=0.40\columnwidth]{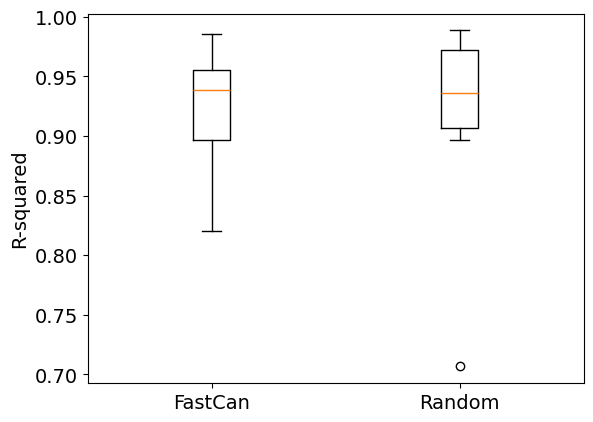}
      \label{fig:box_dsed_opt}%
    }
    \hspace{1cm}
    \subfloat[WHS dataset]{%
      \includegraphics[width=0.40\columnwidth]{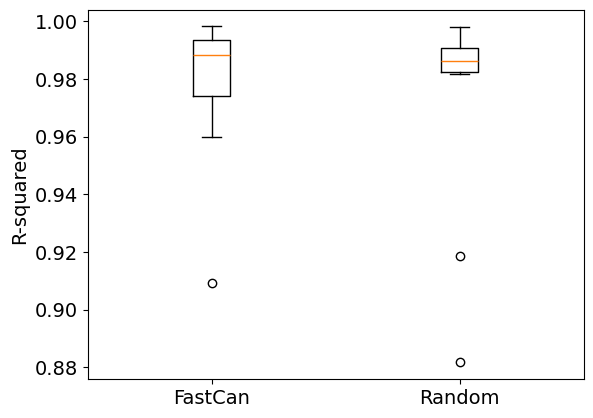}
      \label{fig:box_whbm_opt}%
    }
    \caption{Comparison of sample selection performance between the mini-batch FastCan method with tuned atom size and random selection using R-squared on two datasets}\label{fig:box_three_opt}
\end{figure}

In Fig. \ref{fig:batch_four}, when changing the batch size for each dataset, the number of atoms is fixed at the optimal atom size obtained from Fig. \ref{fig:atom_four}, where the optimal atom sizes $q$ for the SDSE, ADSE, EMPS, and WHS datasets are 15, 20, 25, and 5, respectively.
According to Step 2 in Algorithm \ref{alg:mini_batch_DP}, the maximum batch sizes for the DSE, EMPS, and WHS datasets are 7, 5, 4 and 10, respectively.
It can be seen that when the atom size is fixed at the optimal value, using a larger batch size to reduce redundancy can generally improve the performance of the selected samples. 
Therefore, the atom sizes and the batch sizes for the previous four case studies are the optimal atom sizes obtained from Fig. \ref{fig:atom_four} and the corresponding maximum batch sizes. 
However, it is worth noting that in Fig. \ref{fig:batch_four}, the relationship between R-squared and batch size is not a straightforward linear relationship, especially for the EMPS and WHS datasets. 
A more sophisticated method can be developed in the future to determine the appropriate hyperparameters for the proposed method. 

To examine the underlying causes of the previously observed exceptions—where random selection outperforms or matches the performance of the mini-batch FastCan method, as shown in Figs. \ref{fig:errbar_sdsed} and \ref{fig:errbar_whbm}—Fig. \ref{fig:sample_four} presents the impact of atom size on model performance across varying selected-sample sizes for the mini-batch FastCan approach. This analysis aims to assess whether suboptimal hyperparameter settings contribute to these discrepancies.

To investigate the potential causes for the previously noted exceptions where random selection outperforms or matches mini-batch FastCan selection in Figs. \ref{fig:errbar_sdsed} and \ref{fig:errbar_whbm}, the effect of atom size on model performance corresponding to the mini-batch FastCan method across different selected-sample sizes is demonstrated in Fig. \ref{fig:sample_four}. For each sample size, the maximum R-squared value is printed out.
It can be observed that when the mini-batch FastCan method performs close to or worse than the random method, the adopted atom size is not the optimal size for the given number of selected samples.
For the ADSE dataset, when 60 samples are selected, the optimal atom size is 5 rather than 15. 
For the WHS dataset, when 40 samples are selected, the optimal atom size is 10 rather than 5. 
After tuning the atom size for the mini-batch FastCan method, the comparison of sample selection performance of the two methods is then illustrated in Fig. \ref{fig:box_three_opt}. The results indicate that the mini-batch FastCan method can provide better sample selection results when the appropriate atom size is applied. 

In addition, as shown in Figs. \ref{fig:sample_dsed}, \ref{fig:sample_emps} and \ref{fig:sample_whbm}, when the number of selected samples exceeds a certain threshold, the optimal atom size no longer changes with the number of selected samples.
At the same time, Figs. \ref{fig:errbar_sdsed}, \ref{fig:errbar_emps} and \ref{fig:errbar_whbm} indicate that the performance of the selected samples also stabilises when the number of selected samples exceeds this threshold for each dataset. 
This consistent phenomenon suggests that when more samples cannot contribute additional information, the number of optimal atoms learned from the selected samples tends to stabilise. 
This finding supports, to a certain extent, the rationality and robustness of using dictionary learning to generate pseudo labels. 
Moreover, careful tuning of hyperparameters is particularly important when the number of selected samples is small.

\section{Conclusions}
This paper introduces a dictionary learning-based data pruning method, called mini-batch FastCan, for selecting informative time-series samples in the identification of discrete systems with finite orders.
Specifically, k-means-based dictionary learning is well-suited for generating pseudo-labels in the data pruning process, mitigating the impact of imbalanced candidate sample distributions on sample selection.
Additionally, selecting samples in separate batches allows for the selection of more samples than the number of necessary model terms, while effectively accounting for redundancy within each batch. This method helps ensure that the selected samples are more diverse and informative.

The selection process is independent of the specific system identification technique used, which enhances its versatility.
Although hyperparameters, including atom size and batch size, can be empirically determined by engineering practices, the development of theoretically grounded optimisation strategies is recommended to improve generalizability and methodological rigour in future work.

The case studies on the synthetic and benchmark datasets show that the proposed method effectively reduces the sample size without significantly degrading the identification performance of the benchmark system. 
Compared to random selection, the data-pruning method generally yields better results, with a higher mean and lower variance of R-squared values.
Additionally, it is found that the careful selection of hyperparameters is crucial, particularly when the proportion of selected samples is small.

\section*{Acknowledgements}
The authors would like to acknowledge the support for this work from Shanghai Qi Zhi Institute Innovation Program (SQZ202310) and National Natural Science Foundation of China (52378187).

For the purpose of open access, the authors have applied a Creative Commons Attribution (CC BY) licence to any arising Author Accepted Manuscript version.

\section*{Appendix 1: Prediction Performance of the Baseline NARX}\label{appx1}
For each of the four datasets, a reduced polynomial NARX model with ten terms and an intercept is derived using the full dataset as a baseline. 
To select relevant nonlinear terms, the maximum input and output lags are set to 4 for the SDSE system, ADSE system, and EMPS, and to 7 for the WHS. 
For all systems, the polynomial degree is fixed at three, and the number of selected model terms is limited to ten. 
These parameter settings are designed to provide reasonable predictive performance, but can be further optimised based on specific application requirements. 

\begin{figure} [htbp]
\centering
    \subfloat[Test 1]{%
      \includegraphics[width=0.38\columnwidth]{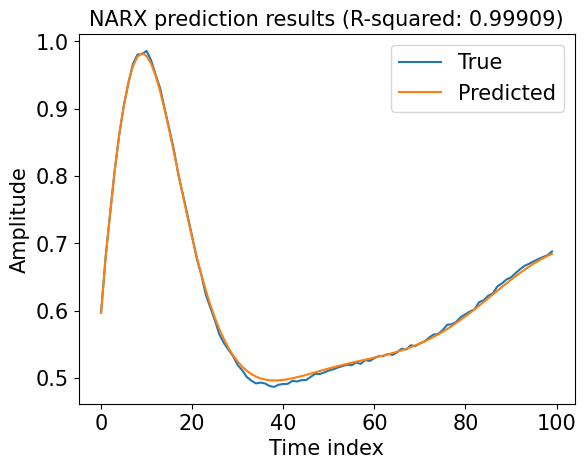}
      \label{fig:pre_test1_sdsed}%
    }
    \hspace{1cm}
    \subfloat[Test 2]{%
      \includegraphics[width=0.40\columnwidth]{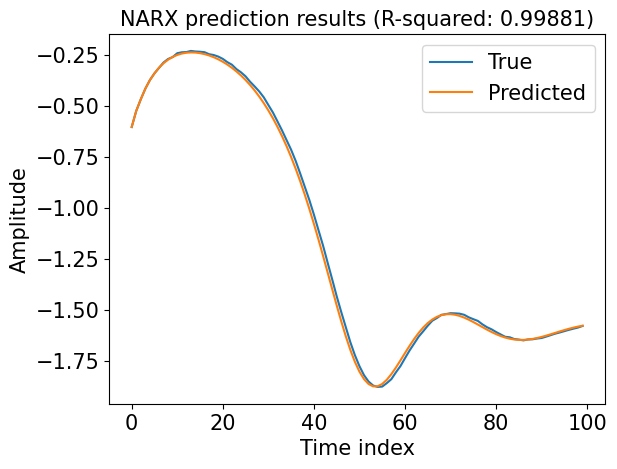}
      \label{fig:pre_test2_sdsed}%
    }
    \caption{The performance of the baseline NARX model for SDSE test data with different initial conditions}\label{fig:pre_sdsed}
\end{figure}

\begin{figure} [htbp]
\centering
    \subfloat[Test 1]{%
      \includegraphics[width=0.38\columnwidth]{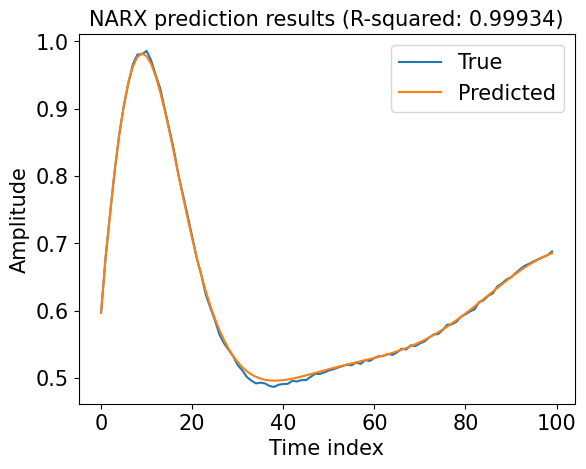}
      \label{fig:pre_test1_adsed}%
    }
    \hspace{1cm}
    \subfloat[Test 2]{%
      \includegraphics[width=0.40\columnwidth]{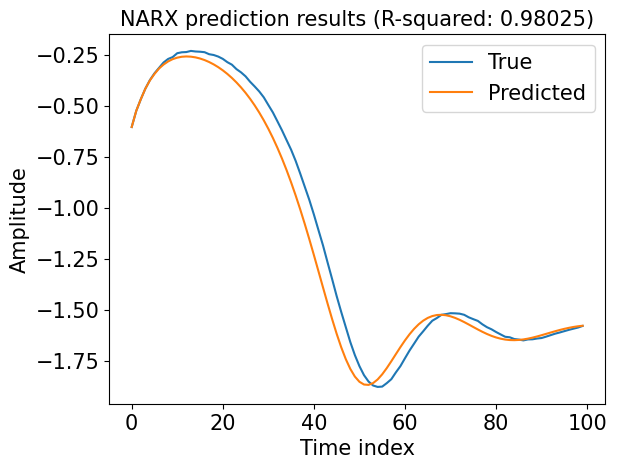}
      \label{fig:pre_test2_adsed}%
    }
    \caption{The performance of the baseline NARX model for ADSE test data with different initial conditions}\label{fig:pre_adsed}
\end{figure}

\begin{figure}[htpb]
\centering
    \subfloat[EMPS dataset]{
    \includegraphics[width=0.40\linewidth]{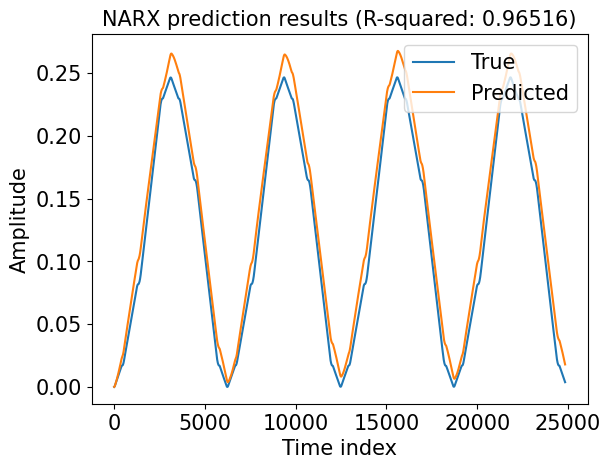}
    \label{fig:pre_emps}
    }
    \hspace{1cm}
    \subfloat[WHS dataset]{
    \includegraphics[width=0.40\linewidth]{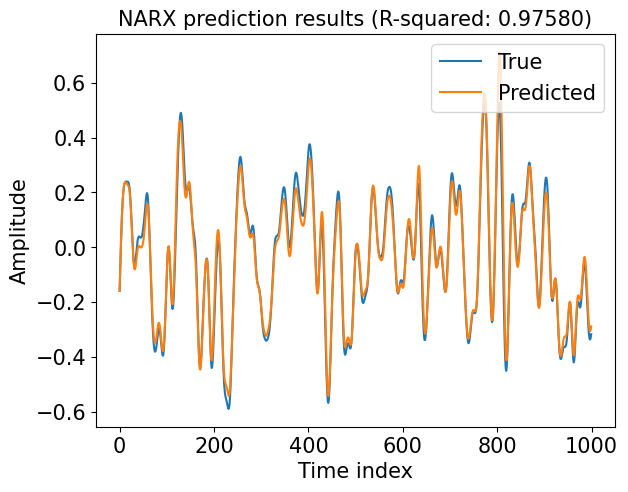}
    \label{fig:pre_whbm}
    }
    \caption{The performance of the baseline NARX model for two benchmark datasets}
    
\end{figure}

\newpage

\bibliographystyle{unsrt}
\bibliography{ref.bib}

\end{document}